\documentclass[pdflatex,sn-mathphys-num]{sn-jnl}% Math and Physical Sciences Numbered Reference Style 
\usepackage{graphicx}%
\usepackage{multirow}%
\usepackage{amsmath,amssymb,amsfonts}%
\usepackage{amsthm}%
\usepackage{mathrsfs}%
\usepackage[title]{appendix}%
\usepackage{xcolor}%
\usepackage{textcomp}%
\usepackage{manyfoot}%
\usepackage{booktabs}%
\usepackage{algorithm}%
\usepackage{algorithmicx}%
\usepackage{algpseudocode}%
\usepackage{listings}%
\usepackage{pgfplots}
\usepackage{placeins}
\usepackage{fix-cm}
\pgfplotsset{compat=1.17}

\theoremstyle{plain}%
\begin{document}

\title[AI-Managed Emergency Documentation with a Pretrained Model]{AI-Managed Emergency Documentation with a Pretrained Model}


\author[1,2]{\fnm{David} \sur{Menzies}} \email{david.menzies@ucd.ie}

\author[2]{\fnm{Sean} \sur{Kirwan}} \email{sean@medwrite.ai}

\author[3]{\fnm{Ahmad} \sur{Albarqawi}} \email{ahmada8@illinois.edu}

\affil[1]{\orgname{School of Medicine, University College}, \orgaddress{\state{Dublin}, \country{Ireland}}}

\affil[2]{\orgname{MedWrite}, \orgaddress{\state{Dublin}, \country{Ireland}}}

\affil[3]{\orgname{University of Illinois Urbana-Champaign}, \orgaddress{\state{Illinois}, \country{USA}}}


\abstract{This study investigates the use of a large language model (LLM)-based system to improve efficiency and quality in Emergency Department (ED) discharge letter writing. Time constraints and infrastructural deficits make compliance with current discharge letter targets difficult. We explored potential efficiencies from an Artificial Intelligence (AI) software model in the generation of ED discharge letters and the attitudes of doctors toward this technology. The evaluated LLM system leverages advanced techniques to fine-tune a model to generate discharge summaries from short-hand inputs, including voice, text, and Electronic Health Record (EHR) data. Nineteen physicians with emergency medicine experience evaluated the system’s text and voice-to-text interfaces against manual typing. The results showed significant time savings with MedWrite LLM interfaces compared to manual methods.}

\keywords{large language models, natural language processing, discharge letters, emergency department}



\maketitle

\section{Introduction}\label{sec1}

Effective communication with General Practitioners (GPs) is crucial for ensuring seamless, safe continuity of care for patients discharged from Emergency Departments (EDs). However, this aspect of healthcare delivery has been identified as a risk area, characterized by poor compliance and challenges related to the content, timeliness, and legibility of discharge letters.

Recently, the integration of artificial intelligence (AI) has revolutionised various facets of medical practice, offering innovative solutions to complex challenges. In the context of EDs, where there are combined pressures for both efficiency and accuracy, AI has the potential to streamline documentation processes and also enhance patient care outcomes.

One such aspect of ED workflow is the composition of discharge letters, which are important communication tools summarizing a patient’s visit, diagnoses, treatments, and follow-up instructions for other healthcare providers (such as their General Practitioner) and potentially also for the patient themselves. The writing of ED discharge letters represents a significant burden on healthcare providers in the ED, requiring careful attention to detail while simultaneously managing other cases. Manual composition of these letters can be time-consuming, susceptible to errors, and variable between healthcare providers. This can lead to delays in communication and difficulties in ensuring appropriate follow-up care.

GP letters are an important means of sharing clinical information between the hospital system and a patient’s primary care provider. These letters should be issued in a timely fashion and contain pertinent information regarding a patient’s care \textsuperscript{\citep{Discharge-to-GP}}. Such communication is also often a regulatory requirement on the discharging doctor \textsuperscript{\citep{Guide-Medical}}.

Compliance with both the timeliness and content of such communications is known to be poor \textsuperscript{\citep{Kripalani_2007}}. Adverse events are recognized as a problem post-patient discharge as well as in the hospital \textsuperscript{\citep{Forster345}}. The transfer of information between hospital and outpatient settings has been recognized as a potential contributory factor to adverse events \textsuperscript{\citep{Medical-Errors-2023}}. Multiple risk factors have been identified in association with discharge letters and the potential impact on patient care \textsuperscript{\citep{schwarz-2019}}.

There is a growing interest in exploring the potential efficiencies offered by AI software models in healthcare, including in the generation of ED discharge documentation. Given these challenges, we hypothesized that an AI solution could mitigate the burden associated with ED discharge letter writing while simultaneously improving the efficiency and quality of documentation. The rationale behind this hypothesis lies in the capabilities of AI to analyse data, recognize patterns, and generate appropriate content.

The potential of generative AI to assist with medical tasks has been explored in several areas, including patient-friendly discharge \textsuperscript{\citep{patient-friendly-2024}} and informed consent \textsuperscript{\citep{Informed-Medical-Consent}}.

Several studies have explored the usage of pre-trained language models for biomedical and clinical tasks, including comparison of various language models and demonstration of the effectiveness of fine-tuning for domain-specific applications like biomedicine \textsuperscript{\citep{lewis-etal-2020-pretrained}}. Fine-tuning provided significant enhancement to biomedical tasks, in addition to clinical tasks.

The challenges of obtaining labelled medical data due to privacy regulations are recognised by Chintagunta et al. \textsuperscript{\citep{chintagunta-etal-2021-medically}}. Combining human-labelled and synthetic data generated by a model for medical dialogue summarization using GPT-3 has been demonstrated to be effective. The use of natural language processing and machine learning has been described before in the generation of ICD-10 codes from HER data, but up to now, the medical literature has not described its use in the generation of an Emergency Department discharge letter \textsuperscript{\citep{sammani-2021}}.

The use of ChatGPT prompt engineering to generate letters from a plastic surgery skin cancer clinic has been reported using the standard ChatGPT interface and without a continuous improvement cycle or data protection step \textsuperscript{\citep{ali2023using}}.

Large language models, specifically GPT-3.5 and ChatGPT, have been evaluated in the domain of medical evidence summarization \textsuperscript{\citep{tang-2023}}. Although this work reveals promising results in terms of the models' ability to generate coherent and comprehensive summaries, challenges are also identified as these models are prone to generating factually inconsistent summaries. 

It was anticipated that MedWrite’s intelligent assistance would reduce the time required to complete discharge letters, streamline workflow processes, and enhance the overall quality and consistency of documentation by incorporating an interface for physicians to easily report and rectify inaccuracies. Through this study, we aimed to evaluate the efficacy of MedWrite in addressing the challenges of ED discharge letter writing, thereby contributing to the advancement of AI-driven solutions in healthcare documentation.

\section{System Development}\label{sec2}

The system core component is GPT-3 Davinci model \textsuperscript{\citep{brown2020language}}, fine-tuned to specialize in the medical writing domain. This fine-tuning process enables the model to create compliant discharge letters tailored to the healthcare systems in Ireland and the UK without requiring lengthy prompting.

The training data was created with the help of third-party medical writers to generate synthesized patients’ medical cases for the emergency department and write the related discharge letters; then, the letters will be processed to add metadata for the training process, such as the target department and country writing style.

The system also uses the open-source whisper model \textsuperscript{\citep{radford2022robust}} for accurate speech-to-text dictation. The model is prompted with medical context and terminology to ensure precise speech recognition capabilities, and the prompt is tuned over multiple iterations to eliminate common errors in translating some medical terms.

\subsection{System Architecture} \label{sec21}

The architecture integrates three key components for efficient data flow and improved medical discharge letter generation:

\begin{itemize}
  \item \textbf{User Interface}: This component enables healthcare professionals to create medical discharge letters and report any hallucinations (fabricated information generated by the AI model) for continuous improvement.
  \item \textbf{Data Processing Pipeline}: This layer strips any personally identifiable information from the doctor's notes to ensure compliance with data protection standards, and it also adds relevant metadata for the training process.
  \item \textbf{Model Fine-Tuning}: Scheduled fine-tuning task is triggered to improve the model's performance based on the latest training data by physicians. For the study, we used 144 discharge letters submitted by doctors and 100 synthesized letters generated by the GPT-3 model from the same training data distribution.
\end{itemize}

\begin{figure}[ht]
\centering
\includegraphics[width=0.95\textwidth]{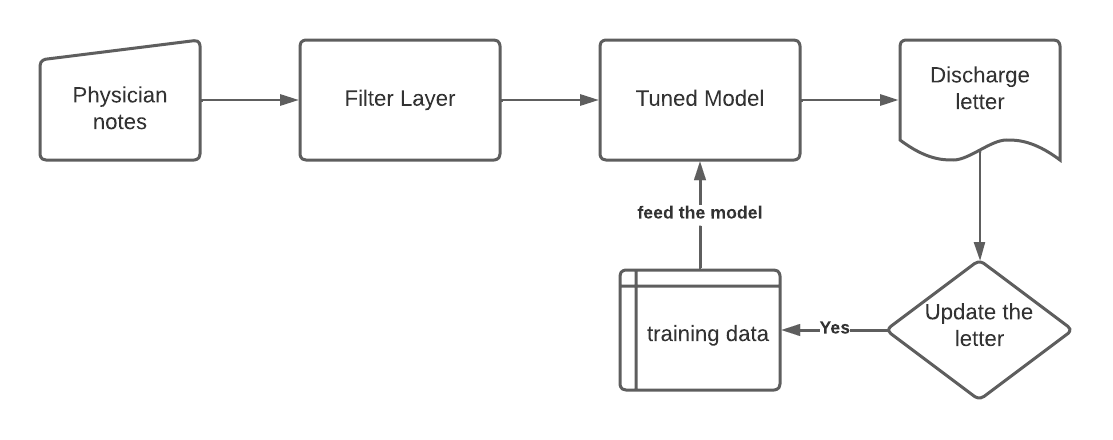}
\caption{High level system design of medical discharge letter generation system highlighting the data flow.}
\label{fig:meddiagram}
\end{figure}

The system design (Figure \ref{fig:meddiagram}) provides a continuous improvement cycle for the generation of discharge letters. The physician will use the interface to input patient visit records, brief diagnoses, and follow-up notes. These data go through a filtration layer to remove patient identity details before sending it to the tuned model, protecting the user information from being mixed with the model training data and enhancing privacy.

The fine-tuned model generates the discharge letter from the anonymized data and reference the notes from the physician’s input. Physicians, with appropriate permissions, can review the generated letter, address any inaccuracies (”hallucinations”), and store the corrected version back into the training data. In the subsequent training cycle, all the data is used to re-tune the Davinci model from scratch. This approach avoids the potential for the model to be biased toward the latest data and ensures a comprehensive model output. This flow allows for direct collaboration with the physicians to tune the model; before the study started, we froze the training pipeline for fair baseline comparison.

\subsection{Data and Evaluation} \label{sec2.2}

The fine-tuning process leveraged 244 discharge letters for the training dataset, containing a simulated patient scenario tailored to the emergency department. This data includes 144 records crafted by five physicians and 100 records of synthesised letters by using synonyms and minor alternations of the existing format. The patient record to generate the letters included vital signs, blood results, treatment plan, general notes, diagnosis, specialist consultation, x-ray / imaging results, and patient history.

The model was trained on the discharge letters in two input formats: one incorporating detailed visit records that mimicked electronic health records (EHR), and another using a concise summary. This approach aimed to ensure the fine-tuned model was robust in generating letters from both comprehensive and short notes. To protect patient privacy and ensure that the model does not expose or mix patient data, we removed personally identifiable information (PII) by replacing sensitive details with placeholders.

Following the initial training, physicians evaluated the model’s output and reported common issues and mistakes in the generated letters, such as referencing the wrong blood result value. This feedback is used to correct the generator letter and then retrain the model again using existing data and fixed letters. The fine-tuning went through four iterations before the study to include the letters with the fixes.

To ensure that the model could adapt to regional regulations and medical depart-mental, the discharge letters were tagged with metadata including the target country (Ireland, UK, or a global) and the department (emergency medicine in this case). This approach enabled the tuning of a single model capable of generating discharge letters that complied with specific regulations and department practices.

\section{Methods}\label{sec3}

An unblinded cohort methodology was used to test both the hypothesis that Med-Write would improve the efficiency of ED discharge letter-writing, and to evaluate the software’s user interface (UI).

Nineteen physicians with at least six months experience working in Emergency Medicine participated in the study (Table~\ref{tab:grade_experience_interest}). Recruitment was via email to an emailing list of Emergency Medicine groups and via social media.

Pre-participation information regarding the MedWrite model was provided to participants. The MedWrite model was also demonstrated to participants online as part of the study, and all participants were provided with an opportunity to practice generating a discharge letter.

Two of the authors (D. Menzies and S. Kirwan) conducted the interviewing and testing over Microsoft Teams, facilitating accurate timing of each part of the process. Each participant undertook three timed tasks to generate an ED discharge letter based on a fictional EHR: manual composition (typing), AI text input, and AI voice-to-text interface. 

In the case of the manual composition, participants were asked to write a letter to their usual work practice standard. In the case of the AI-assisted letters, participants were asked to edit the AI output until it met their usual standard. In addition to timed tasks, the participants’ experiences and attitudes towards Emergency Medicine Discharge Letter-Writing, MedWrite and their experience with the software were explored through a semi-structured interview.

All participants were provided with a Participant Information Leaflet in advance of the study and were given an opportunity to ask any clarifying questions. All participants opted into participation and all consented to participation in the study prior to commencing.

\clearpage
\section{Results}\label{sec4}

\begin{table}[h!]
\begin{tabular}{|l|c|c|}
\hline
\textbf{Input Method} & \textbf{Mean Time (seconds)} & \textbf{Range (seconds)} \\
\hline
Manual Typing & 196.737 & 86 - 355 \\
\hline
Using AI Text Interface & 94.58 & 34 - 147 \\
\hline
Using AI Voice to Text & 137.933 & 86 - 208 \\
\hline
\end{tabular}
\caption{Compare the discharge letter completion time using three methods.}
\label{tab:discharge_letter}
\end{table}

\begin{figure}[h!]
\centering
\includegraphics[width=0.95\textwidth]{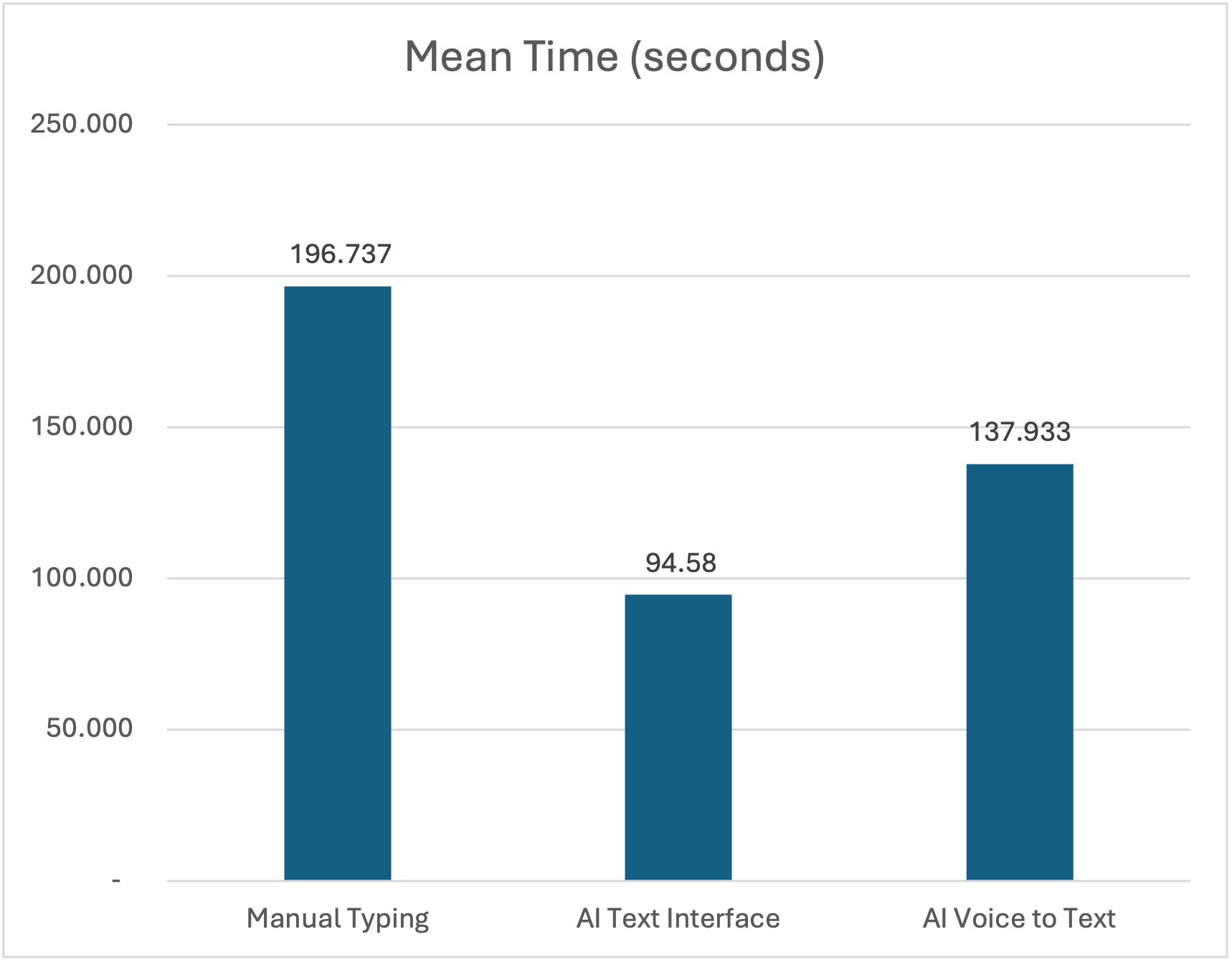}
\caption{Compare the average completion times for discharge letters using manual typing and AI-assisted generation. The AI-assisted generation uses simulated EHR input using text-based interface or voice dictation.}
\label{fig:discharge_letter_times}
\end{figure}

The time taken to generate discharge letters was measured as shown in  Table~\ref{tab:discharge_letter}. Participants took a mean time of 196.737 seconds (range 86-355 seconds) to complete a manual (typed) discharge letter. 

The AI text interface is a simulation of using an EHR system where the user copies the patient details from it and generates the letters using the LLM, it took a mean time of 94.58 seconds (range 34-147 seconds), representing a mean time savings of 102.157 seconds. This equates to an average time saving of 51.9\%. 

The AI voice-to-text interface took a mean time of 137.933 seconds (range 86-208 seconds), with a mean time savings of 58.804 seconds. This equates to an average time saving of 37\%. The measured time for AI-assisted generation also accounts for manual corrections to the generated letters before completion. 

During the semi-structured interviews, 18 participants (94\%) identified several ‘pinch points’ in EDs, including negative experiences with IT. 11 participants (58\%) reported not completing all GP discharge letters.

Participants estimated that they completed an average of 11 discharge letters per shift. Using the mean recorded time to generate a manual discharge letter (196.737 seconds) in this study, that equates to an average manual letter writing time by participant per shift of 36 min 4 sec (2,164.107 seconds). If the AI Text Interface was used by these emergency medicine doctors in a shift with the same average discharge letter volume, it would take 17 min 20 sec (1040.37 seconds) per shift, per doctor. This would provide time savings of 18 minutes 44 seconds per shift, per doctor. 

\ \\
\textbf{Calculations of average reported writing per shift, per doctor vs an expected AI time savings}

\begin{equation}
\begin{aligned}
\text{Average Manual Writing (sec)} &= 11 (letter) \times 196.737 (sec) = 2,164.107 \\
\text{Average AI Interface Writing (sec)} &= 11 (letter) \times 94.58 (sec) = 1,040.38
\end{aligned}
\end{equation}

\section{Discussion}\label{sec5}

This study investigated the efficiency of using a large language models-based system to improve the efficiency of emergency department (ED) discharge letter writing, requires minimal editing of the generated letter instead of manually writing it from scratch.

The findings support the proposition that the MedWrite fine-tuned model can reduce completion time compared to manual typing, 15 of the participants indicated they would try where it is available, and 12 indicated they would definitely use it. Voice dictation was preferred over manual note writing. Both text-based and voice-to-text interfaces resulted in substantial time savings and improved writing quality.

Participants identified several administrative burdens in the emergency department, including access to Information and Communication Technology (ICT), which impacted their ability to complete discharge letters. Most participants also reported that a product such as MedWrite would reduce these burdens and improve compliance.

The Royal College of Emergency Medicine recognises that AI has a potential benefit in Emergency Medicine and provides some guidance on how it should be implemented \textsuperscript{\citep{AI_Position_Statement}}, however this focuses largely on the use of AI as a decision or diagnosis support tool rather than automating administrative tasks.

\section{Limitations}\label{sec6}

Despite promising results, some limitations are present in this study. The study was conducted in a simulated environment and applying the system in a real-world hospital environment is necessary to assess the impact of LLMs on workflow efficiency and user satisfaction. 

The number of participants in the study was small, intended to demonstrate a ‘proof of concept’ of MedWrite, rather than reach statistical significance. The study interviews, and timed tasks, were administered by two of the authors – neither authors nor participants were blinded to the process, thus risking the introduction of bias.

Each participant completed a practice letter and the three timed letters. Different content was used for each and differences in complexity may have impacted the results. In particular, the voice to text EHR provided, was intentionally briefer as it was intended to test user satisfaction with the interface as much as the potential time saving versus the other input methods. In all cases however, participants were required to edit the outputs until they were satisfied with the quality of the letters, which should act as an internal control.

The model’s susceptibility to generating inaccurate information (”hallucinations”) in some scenarios requires continuous model training, which we managed to reduce by incorporating the evaluator feedback in the training data. However, such feedback was paused during the study time period in order to ensure consistency between participants.

Integration with the EHR was simulated in this study. True benefits of the software may be better or worse than those demonstrated when a full EHR integration is implemented.

\section{Conclusion}\label{sec7}

In conclusion, emergency department discharge letters are recognized as an important part of communication regarding patient care, ensuring continuity of care, post-discharge instructions, and patient education. Compliance with the timeliness and content of these letters is variable.

Generative AI, particularly with trained LLMs, has the potential to add efficiencies to labour-intensive administrative tasks in healthcare. We hypothesized that a trained LLM could accurately generate plain English discharge letters using a variety of inputs including electronic healthcare records, free text, and speech. We fine-tuned the GPT3 Davinci model which helped tailor the discharge letter writing style and incorporate the feedback of physicians.

We tested this hypothesis in a medical interface MedWrite among 19 emergency medicine doctors. This study suggests that AI-assisted systems have the potential to reduce the time taken to generate suitable discharge letters while retaining comparable levels of content.

\section{Declarations}\label{sec8}

MedWrite system received funding from the Wicklow Local Enterprise office.

Ethical approval for the study was obtained from the University College Dublin, Human Research Ethics Committee (UCD HREC) with reference number: LS-C-23-116-Menzies.

\section{Data Availability}\label{sec9}

The data supporting the findings of this study is available on request.

\section{Author Contributions}\label{sec10}

All authors contributed equally to the writing of the manuscript. 
David Menzies (DM) was the main driver of the study interviews. Sean Kirwan (SK) compiled the study results and participated in the interviews. Ahmad Albarqawi (AA) fine-tuned the language model and developed the system interface. 

\section{Appendix}\label{sec11}

\begin{table}[h!]
\centering
\begin{tabular}{|c|l|c|c|l|}
\hline
\textbf{Index} & \textbf{Grade}              & \textbf{Experience in ED Years} \\ \hline
1              & General Practitioner        & 1                             \\ \hline
2              & Consultant                  & 9   \\ \hline
3              & Medical Officer Year 2      & 1.08    \\ \hline
4              & Registrar                   & 4      \\ \hline
5              & SHO                         & 3   \\ \hline
6              & Consultant                  & 13   \\ \hline
7              & Registrar                   & 3   \\ \hline
8              & Senior Registrar            & 5    \\ \hline
9              & Recent Consultant           & 10  \\ \hline
10             & Registrar                   & 9   \\ \hline
11             & Consultant                  & 12  \\ \hline
12             & Consultant                  & 20   \\ \hline
13             & SHO                         & 3   \\ \hline
14             & Consultant                  & 34   \\ \hline
15             & Registrar                   & 3     \\ \hline
16             & SHO                         & 1.5    \\ \hline
17             & Recent Consultant           & 8    \\ \hline
18             & Senior Registrar            & 6    \\ \hline
19             & Specialist                  & 6    \\ \hline
\end{tabular}
\caption{Grade and experience of the study participants.}
\label{tab:grade_experience_interest}
\end{table}

\backmatter

\clearpage
\bibliography{sn-bibliography}% common bib file

\end{document}